\documentclass{article}
\usepackage{ijcai25}
\usepackage{wrapfig}
\usepackage{times}
\usepackage{soul}
\usepackage{url}
\usepackage[hidelinks]{hyperref}
\usepackage[utf8]{inputenc}
\usepackage[small]{caption}
\usepackage{graphicx}
\usepackage{amsmath}
\usepackage{amsthm}
\usepackage{amssymb}

\usepackage{booktabs}
\usepackage{algorithm}
\usepackage{algorithmic}
\usepackage[switch]{lineno}

\title{MemeBlip2: A Novel Light Weight Multimodal System to Detect Harmful Memes}

\author{
Jiaqi Liu$^{1}$\and
Ran Tong$^{2}$\and
Aowei Shen$^{1}$\and
Shuzheng Li$^{1}$\and
Changlin Yang$^{1}$\and
Lisha Xu$^{1}$
\affiliations
$^1$Independent Researcher
$^2$Mathematics and Statistics Department, University of Texas at Dallas
\emails
jackyliu9747@gmail.com,
rxt200012@utdallas.edu,
aowei.shen.ml@gmail.com,
lishuzheng1012@gmail.com,
chlinyang97@gmail.com,
xlisha0228@gmail.com
}

\begin{document}

\maketitle

\begin{abstract}
Memes combine images with brief text to convey humour, opinions, and social commentary, but they can also spread harmful content such as hate speech. We present MemeBLIP2, a lightweight multimodal system that detects harmful memes by jointly modelling visual and textual signals. The method extends prior work by incoporating BLIP‑2 vision–language encoders. Our model is evaluated on the PrideMM dataset, where it reaches $77.5\%$ accuracy, $81.8\%$ AUROC, and a $79.0\%$ macro F1 score. The results demonstrate that MemeBLIP2 captures subtle cross‑modal cues, including irony and culturally specific references, and it surpasses existing multimodal baselines on the benchmark.
\end{abstract}

\section{Introduction}

Memes—images combined with short text—have become a popular form of online communication. They mix pictures and words to share humor, satire, or opinions about social and political issues. Because memes spread quickly and widely, they have become an important tool for activism and social commentary. However, memes can sometimes carry harmful messages, such as hate speech or harassment aimed at specific groups or individuals. These harmful memes can promote prejudice, discrimination, and negatively impact online safety and social harmony. Therefore, identifying harmful memes is a growing area of research, driven by a need to address online harm.

Detecting harmful memes requires understanding how pictures and text work together. An important early effort was the Hateful Memes Challenge by Kiela et al.~\cite{Kiela2020}, which provided a dataset specifically designed to test multimodal hate detection. This dataset was challenging because the text or image alone might seem harmless, but combined they became harmful. This highlighted that analyzing images and text separately is insufficient.

Following this challenge, researchers began using powerful computer models trained on large amounts of visual and textual data. For example, the MOMENTA system~\cite{Pramanick2021} not only identified harmful memes but also determined the target of the hate speech by leveraging pre-trained model features. Later, significant improvements came with the CLIP model~\cite{Radford2021}, a system known for effectively matching images with text by learning from vast internet data. Systems like HateCLIPper~\cite{Kumar2022}, built specifically around CLIP, showed notable improvements in detecting hateful memes. Recently, MemeCLIP~\cite{Shah2024} achieved state-of-the-art results by efficiently combining visual and textual information from CLIP, yet subtle meanings and cultural references still posed challenges.

To address these ongoing difficulties, we propose MemeBLIP2, a new lightweight multimodal system. MemeBLIP2 takes inspiration from MemeCLIP but introduces BLIP-2~\cite{Li2023}, a more advanced vision-language model. BLIP-2 better captures subtle details and deeper connections between text and images due to its refined training approach. We expect MemeBLIP2 to significantly improve the detection of hidden or nuanced harmful messages in memes, thus making online spaces safer and more inclusive.

\section{Related Work}
Early efforts on harmful content detection were mostly limited to modalities, i.e., analyzing text or images independently. Some studies tried to detect toxic content by building out hate speech lexicons, or training CNN models to recognize hate symbols in images. But these methods found it difficult to account for the complexity of memes, where harm oftencomes from an ironic mismatch of text and visuals, such as a “peace” slogan inserted into a violent image. Such cross-modal semantic conflicts were missed by single-modality models, which led to the need for more nuanced methods.

Then came the Hateful Memes Challenge \cite{Kiela2020}, a pivotal moment in which meme analysis was framed as a multimodal task. This benchmark showed humans outperformed single-modality AI systems by a large margin, motivating the need for joint text-image understanding. Later work harnessed pretrained vision-language models (VLMs) such as CLIP to fill the gap. For instance, MOMENTA \cite{Pramanick2021} combined CLIP embeddings with social context graphs to improve detection accuracy, and HateCLIPper \cite{Kumar2022} employed adversarial training to alleviate CLIP’s over-reliance on textual cues. These techniques made significant advances, reporting up to 65–75\% accuracy on datasets like the Facebook Hateful Memes dataset. But a key limitations remain. The CLIP-based models, designed to maximize global text-image similarity, commonly failed to connect the fine-grained region-wise interactivity properties (e.g., biased symbols in image corners with neutral captions). Performance gaps also occurred in culturally specific contexts, where sarcastic memes that portray minority groups (e.g., LGBTQ + communities) were often misclassified.

Recent work has attempted to address these issues with disentangled feature learning and community-centric datasets. For instance, have decoupled CLIP's text and image encoders to improve generalization \cite{Shah2024}, and datasets such as PrideMM concentrated on underrepresented cultural contexts. Our new routines may be more advanced, but we still have core problems. Static CLIP encoders are ill-suited to the fast-paced evolution of meme styles, including AI-generated visuals, or stylized text. Region-aware methods, which are representative of external object detectors (i.e.YOLO), introduce another level of computational complexity, whilst the opaque nature of the decision-making processes — especially in memes containing sarcasm or cultural metaphors — are a limitation on their practical application for human moderators.

BLIP\cite{Li2022} and its successor BLIP2\cite{Li2023} are unified vision-language models based on the multimodal mixture-of-encoder-decoder (MED) architecture, supporting both understanding and generation tasks. A core component in both models is the image-grounded text encoder, which integrates visual information into textual representations via cross-attention layers inserted between the self-attention and feed-forward modules in each transformer block. A task-specific [Encode] token is added to the input text, and its output embedding serves as the joint image-text representation. Trained with an image-text matching (ITM) loss, this encoder enables fine-grained alignment between modalities, facilitating strong performance in tasks such as image-text retrieval, image captioning, and visual question answering\cite{Li2022}.

BLIP2\cite{Li2023} builds on this foundation with enhanced efficiency, scalability, and generative capabilities. It introduces parameter-efficient fine-tuning through lightweight adapter modules, significantly reducing the number of trainable parameters without compromising performance. With diverse training objectives and improved architectural designs, BLIP2 achieves better alignment and more fluent, context-relevant generation. Furthermore, its pre-training on a broader and more diverse image-text dataset strengthens its generalization across domains.

These advancements make BLIP2 particularly suitable for multimodal hate speech detection, where understanding nuanced visual and textual signals is crucial. Our BLIP-2-based approach is expected to provide significant improvements over prior methods by effectively addressing region-specific semantics and subtle inter-modal mismatches common in memes. The fine-grained modality alignment, enabled by the image-grounded encoder, facilitates enhanced understanding of sarcasm, cultural metaphors, and rapidly evolving meme styles. Additionally, leveraging BLIP-2’s parameter-efficient adapters ensures efficient adaptation to new meme contexts, thus improving robustness and reducing misclassification rates, particularly in culturally nuanced memes. Its image-grounded encoder, robust generative quality, and efficient fine-tuning collectively enable reliable identification of harmful content in complex multimodal scenarios.

\section{Methodology of MemeBLIP2}
\begin{figure*}[ht]
    \centering
    \includegraphics[width=1\textwidth]{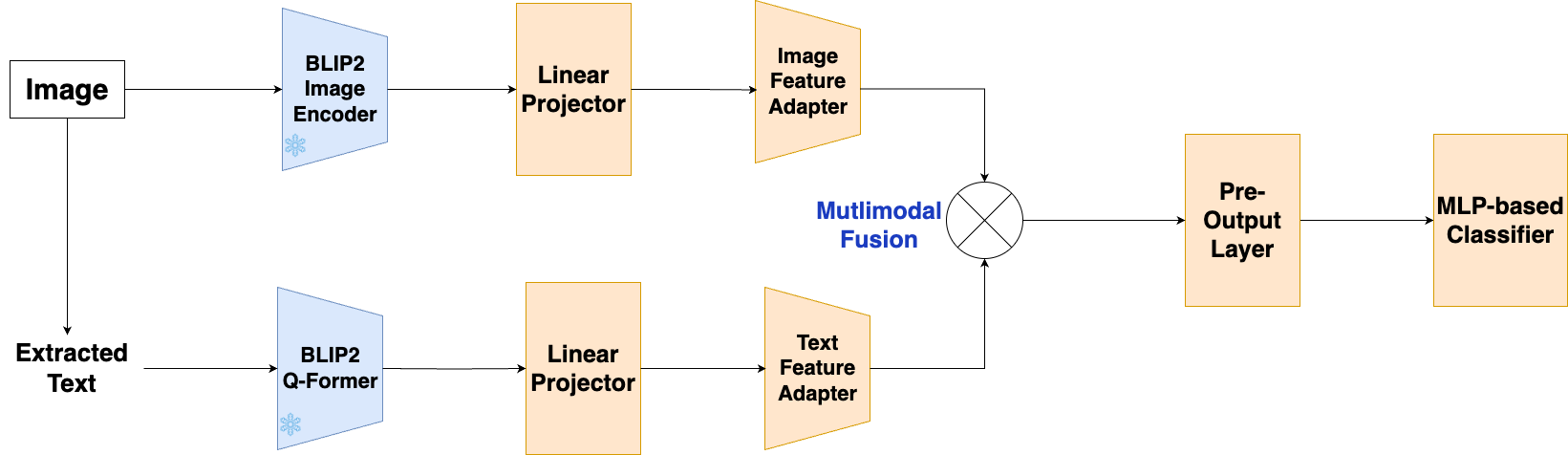}
    \caption{
    An overview of our proposed framework, MemeBlip2. We use frozen Q former image encoder and text encoders from BLIP-2 to create representations for each image-text pair. These representations are passed through linear layers to disentangle the modalities in BLIP-2’s shared embedding space. We implement Feature Adapters with residual connections for each modality to prevent overfitting. We use a MLP-based classifier to make MemeBLIP2 more robust to imbalanced data. We initialize MLP-based classifier weights by using Semantic-Aware Initialization to further improve performance.
    }
    \label{fig:memeblip2}
\end{figure*}

In this section, we describe our proposed framework, MemeBLIP2, for harmful meme classification. 
We provide the detailed mathematical formulation of each component.
The detailed flow of the MemeBLIP2 methodology is outlined in Figure ~\ref{fig:memeblip2}.
\subsection{ Dataset }
We conduct our experiments on the PrideMM dataset~\cite{Shah2024}, a benchmark curated for evaluating multimodal models in the context of social media memes. PrideMM contains over 7,000 memes collected from platforms such as Reddit and Twitter, with each sample consisting of an image and a textual caption. The dataset is annotated for harmfulness and offensiveness, posing a challenging binary classification task.

Unlike traditional vision-language datasets, PrideMM includes a significant number of sarcastic, ironic, or culturally nuanced memes, which require the model to jointly reason over visual and textual content to make correct predictions. The dataset is split into training, validation, and test sets following the protocol established in~\cite{Shah2024}, ensuring fair comparison with prior work such as MemeCLIP.

We start with the careful preparation of the  dataset by loading all images and their matching textual annotations from the provided dataset files. Every image is resized to a consistent resolution of 224x224 pixels. This resizing helps keep the inputs uniform, ensuring the model trains smoothly and efficiently. Labels indicating whether memes are harmful or benign are directly taken from the dataset annotations.

We express the problem as: 
Let $\{(\mathbf{I}_i, t_i, y_i)\}_{i=1}^N$ represent the collection of $N$ samples in the dataset, where $\mathbf{I}_i$ is the $i$-th image, $t_i$ is the associated text, and $y_i \in \{0, 1\}$ is its label, where 0  represents benign content and 1 for harmful content.

\subsection{Linear Projection}
After preparing the datasets, the next step involves extracting meaningful features from both visual and textual data using the BLIP-2 model (Salesforce/blip2-opt-2.7b)~\cite{Li2023}. The BLIP-2 vision encoder produces visual embeddings, which are 1408-dimensional vectors capturing detailed visual characteristics of each meme image by mapping the i the image $\mathbf{I}_i$ to a high-dimensional vector $\mathbf{v}_i \in \mathbb{R}^{1408}$.
\[
\mathbf{v}_i = E_{\text{vision}}({I}_i)
\]

At the same time, the BLIP-2 language encoder (also known as Q-Former) generates textual embeddings, which are 768-dimensional vectors encapsulating the meaning of the associated text, by mapping $t_i$ to $\mathbf{u}_i \in \mathbb{R}^{768}$ 
\[
\mathbf{u}_i = E_{\text{text}}(t_i).
\]
Here, $E_{\text{vision}}(\cdot)$ and $E_{\text{text}}(\cdot)$ denote the pretrained encoders for images and text, respectively.

Although $\mathbf{v}_i \in \mathbb{R}^{1408}$ and $\mathbf{u}_i \in \mathbb{R}^{768}$ are informative, they live in different embedding spaces.
Once we have these embeddings, we can transform them into a shared $1024$-dimensional  space. To achieve this, we use linear projection layers that reduce the visual embeddings from 1408 dimensions and increase textual embeddings from 768 dimensions to a common space of 1024 dimensions. In our model, let \( x \in \mathbb{R}^{d_{\text{in}}} \) be the input embedding vector(from either the vision or language encoder). The first linear projection is given by
\[
x^{(0)} = \operatorname{Dropout}\Big(\operatorname{ReLU}\big(\operatorname{LayerNorm}(W_0 x + b_0)\big)\Big),
\]
where \( W_0 \in \mathbb{R}^{d_{\text{out}} \times d_{\text{in}}} \) and \( b_0 \in \mathbb{R}^{d_{\text{out}}} \) are the weight matrix and bias of the initial linear layer, and \(\operatorname{Dropout}(\cdot)\) applies a dropout with probability \( p \). For additional layers (if the number of mapping layers is greater than one), a residual connection is introduced. Specifically, for \( l = 1, \dots, L-1 \), the transformation is defined as
\[
x^{(l)} = \operatorname{Dropout}\Big(\operatorname{ReLU}\big(\operatorname{LayerNorm}(W_l x^{(l-1)} + b_l)\big)\Big) + x^{(l-1)},
\]
where \( W_l \in \mathbb{R}^{d_{\text{out}} \times d_{\text{out}}} \) and \( b_l \in \mathbb{R}^{d_{\text{out}}} \) are the weights and biases of the \( l \)-th linear layer. The final output \( x^{(L-1)} \) is the projected embedding in the shared latent space denoted as  $\mathbb{R}^{1024}$, 
 ensuring that the features are normalized and enhanced by the nonlinear activation while preserving the original information.


\subsection{Adapter Module.}
Following the projection, an adapter module~\cite{Houlsby2019} further refines the projected features with additional nonlinear transformations. Let $x \in \mathbb{R}^{c}$ be the input to the adapter. First, layer normalization is applied:
\[
\tilde{x} = \operatorname{LayerNorm}(x).
\]
Next, $\tilde{x}$ is passed through a bottleneck that reduces the dimensionality to $\tilde{c} = \max(16, \lfloor c / r \rfloor)$ with reduction factor $r=1.5$, followed by GELU activation~\cite{Hendrycks2016}:
\[
h = W_2\,\operatorname{GELU}\bigl(W_1 \tilde{x}\bigr),
\]
where $W_1\in\mathbb{R}^{\tilde{c}\times c}$ and $W_2\in\mathbb{R}^{c\times \tilde{c}}$ are learned weights. A dropout layer with rate $\rho$ then produces
\[
\hat{h} = \mathrm{Dropout}(h).
\]
We incorporate a scaled residual connection:
\[
x' = x + \alpha\,\hat{h},
\]
where \(\alpha\) is a learnable scalar (initialized to 0.1). This residual connection ensures that the adapter can introduce new nonlinear feature refinements without overwriting the original representation entirely. By scaling the adapter output with \(\alpha\), the model can learn how much of the new information to incorporate.

Finally, a second layer normalization yields the output of the adapter:
\[
y = \operatorname{LayerNorm}(x').
\]
We denote this entire operation as:
\[
\mathrm{Adapter}(x) = y.
\]
 The purpose of the entire adapter module is to introduce additional nonlinear transformations while preserving the original feature information. At the same time,  \(\alpha\) controls the influence of the adapter.
\subsection{Final Adaptation Mix.}
The adapted visual and textual embeddings obtained from the previous two steps are then combined through element-wise multiplication, in order to capture subtle and complex interactions between the image and text. The combined embeddings are normalized to stabilize the training process.And the whole process can be expressed as:

Let $\mathbf{v}_i^\text{proj}$ be the projected image features and $\mathbf{u}_i^\text{proj}$ be the projected text features, which are obtained from the linear projection step. Two adapters introduced in the second step are used, $A_I$ for images adapter and $A_T$ for text adapter, and mix them with a ratio $\beta \in [0,1]$:
\[
\mathbf{v}_i^\text{adapted} 
= \beta \, A_I(\mathbf{v}_i^\text{proj})
+ (1-\beta)\,\mathbf{v}_i^\text{proj},
\]
\[
\mathbf{u}_i^\text{adapted} 
= \beta \, A_T(\mathbf{u}_i^\text{proj})
+ (1-\beta)\,\mathbf{u}_i^\text{proj}.
\]
The results are normalized by :
\[
\mathbf{\bar{v}}_i = \frac{\mathbf{v}_i^\text{adapted}}{\|\mathbf{v}_i^\text{adapted}\|}, 
\quad 
\mathbf{\bar{u}}_i = \frac{\mathbf{u}_i^\text{adapted}}{\|\mathbf{u}_i^\text{adapted}\|}.
\]
\subsection{Multimodal Fusion}
We fuse the refined image and text embeddings using element-wise multiplication:
\[
\mathbf{f}_i = \mathbf{\bar{v}}_i \odot \mathbf{\bar{u}}_i.
\]
This step encodes direct interaction between each dimension of the visual and textual embeddings. A final feed-forward ~\cite{Vaswani2017} block is used to further processes $\mathbf{f}_i$ and generates $\mathbf{z}_i$, known as the fused multimodal embedding:
\[
\mathbf{z}_i = \mathrm{DenseNN}(\mathbf{f}_i),
\]
where $\mathrm{DenseNN}$ is consisted of  a linear layer, normalization, ReLU, and dropout.
\subsection{Classification with MLP-Based Classifier}
After obtaining the fuse multimodal embebdidng $\mathbf{z}_i$, we experimented with two approaches for classifying the fused multimodal embedding $\mathbf{z}_i$: a cosine-based classifier and an MLP-based classifier. Empirically, the MLP classifier yielded higher accuracy in our setup, so we adopt it here as our final design. Specifically, we classify each fused embedding $\mathbf{z}_i$ using layer normalization, a GELU activation, dropout, and a final linear layer. Let $\mathbf{W}\in\mathbb{R}^{K \times d}$ and $\mathbf{b}\in\mathbb{R}^{K}$ be the weight matrix and bias for $K$ classes, where $d$ is the dimension of $\mathbf{z}_i$. The forward pass is as follows:
\[
\begin{aligned}
\tilde{\mathbf{z}}_i &= \mathrm{LayerNorm}(\mathbf{z}_i), \quad \mathbf{h}_i = \mathrm{GELU}(\tilde{\mathbf{z}}_i),\\
\mathbf{h}_i' &= \mathrm{Dropout}(\mathbf{h}_i), \quad \mathrm{logits}_i = \mathbf{W}\,\mathbf{h}_i' + \mathbf{b}.
\end{aligned}
\]

For binary classification ($K=2$), the label is
\[
\hat{y}_i = \arg\max_{k \in \{0,1\}}\,\mathrm{logits}_{i,k}.
\]

Unlike cosine-based methods, which rely on angular similarity, this MLP-based classifier uses learned transformations and normalization to handle scale differences in the embeddings. 
Layer normalization stabilizes training, and the GELU nonlinearity plus dropout relieve the problem of  overfitting. As our empirical results show, these added nonlinear transformations allow the MLP to model subtle semantic cues more effectively, especially in imbalanced datasets, thus outperforming the cosine classifier in our experiments.
\subsection{Training and Optimization}During training, we optimize the classifier parameters using the AdamW optimizer\cite{Loshchilov2018}.
The update rule is defined as:

\begin{equation}
\theta_{t+1} = \theta_t - \eta \left(\frac{\hat{m}_t}{\sqrt{\hat{v}_t} + \epsilon} + \lambda \theta_t\right),
\end{equation}
where \(\theta_{t}\) represents the model parameters at iteration \(t\), \(\eta\) represents the learning rate, \(\hat{m}_{t}\) and \(\hat{v}_{t}\) denote the bias-corrected first and second moment estimates respectively, \(\epsilon\) stands for a small constant included for numerical stability, and \(\lambda\) is the weight decay factor. 


Additionally, we utilize the cross-entropy loss function, which measures the divergence between the predicted class probabilities and the true labels. It is given by:

\[
\text{Cross-Entropy Loss} 
= -\sum_{i=1}^{N} 
   \sum_{k=0}^{1} 
   \mathbf{1}_{\{y_i = k\}} \log(\hat{p}_k(\mathbf{z}_i)),
\]
where $\hat{p}_k(\mathbf{z}_i)$ is the predicted probability for class $k$. 

By combining all the above components---feature projection, adapter refinement, multimodal fusion, and a robust MLP-Based Classifier---MemeBLIP2 is trained end-to-end to  detect harmful memes in our datasets effectively.

\section{Experiments}

\subsection{Evaluation Metrics}
In our experiments, we use three evaluation metrics to assess the performance of MemeBLIP2: Validation Accuracy, Validation AUROC, and Validation F1 Score. These metrics measure different aspects of classification performance and provide a comprehensive assessment of the model's effectiveness.

Validation Accuracy is calculated as the proportion of correctly classified samples out of all samples. Mathematically, it is defined as:
\[
\text{Accuracy} = \frac{TP + TN}{TP + TN + FP + FN},
\]
where \(TP\) represents true positives, \(TN\) true negatives, \(FP\) false positives, and \(FN\) false negatives. Higher accuracy indicates that the model correctly classifies a larger proportion of samples.

Validation AUROC (Area Under the Receiver Operating Characteristic Curve) measures the model's ability to distinguish between classes across various classification thresholds. AUROC is calculated by plotting the True Positive Rate (TPR) against the False Positive Rate (FPR), where:
\[
\text{TPR} = \frac{TP}{TP+FN}, \quad \text{FPR} = \frac{FP}{FP+TN}.
\]
The AUROC is obtained by integrating the TPR over all possible threshold values:
\[
\text{AUROC} = \int_0^1 \text{TPR}(t)\,dt.
\]
An AUROC value close to 1 indicates excellent discrimination between classes, while a value near 0.5 indicates poor performance equivalent to random guessing.

Validation F1 Score is the harmonic mean of precision and recall, providing a balanced evaluation of the model’s performance, especially useful in the presence of imbalanced datasets. Precision measures the proportion of correct positive predictions out of all positive predictions, and recall measures the proportion of correctly identified positives out of all actual positives:
\[
\text{Precision} = \frac{TP}{TP+FP}, \quad \text{Recall} = \frac{TP}{TP+FN}.
\]
The F1 Score combines these metrics as follows:
\[
\text{F1 Score} = 2 \cdot \frac{\text{Precision} \times \text{Recall}}{\text{Precision} + \text{Recall}}.
\]
A higher F1 Score indicates a better balance between precision and recall, demonstrating the model’s effectiveness in identifying positives without producing excessive false predictions.

\subsection{Experimental Setup}

\paragraph{Motivation} 
The design of our experimental framework is grounded in the hypothesis that pre-trained vision-language models (VLMs), such as BLIP-2, encapsulate rich cross-modal semantics that can be effectively adapted to domain-specific tasks through lightweight architectural modifications. Prior works like MemeCLIP~\cite{Shah2024} have demonstrated that preserving frozen pre-trained encoders while fine-tuning minimal projection and adapter modules achieves competitive performance with high efficiency. Our goal is to validate whether similar strategies remain effective when applied to a stronger backbone (BLIP-2 vs. CLIP) on the PrideMM dataset.

\paragraph{Model Configuration} 
We adopt BLIP-2 (Salesforce/blip2-opt-2.7b) as the core backbone, freezing both the image and text encoders to retain general-purpose multimodal understanding. To adapt BLIP-2 representations to our meme classification task, we append modality-specific linear projection layers, lightweight residual adapters, and a shared pre-output transformation. For classification, we use a cosine classifier with learnable bias terms. This setup aligns with recent best practices for low-resource transfer learning, mitigating overfitting while ensuring model flexibility.

\paragraph{Training Details} 
We train for 12 epochs using the AdamW optimizer with a learning rate of $5 \times 10^{-5}$, weight decay of $10^{-4}$, and gradient clipping of 1.0. A warm-up of 3 epochs is followed by cosine annealing learning rate decay. We use mixed precision (fp16) training to improve memory efficiency. The batch size is 64. All experiments are conducted on an NVIDIA GPU with 16GB memory. Validation AUROC is used to select the best model checkpoint.

\paragraph{Evaluation Setup}
To ensure fair and replicable evaluation, we follow the 85\%/5\%/10\% training/validation/test split used in MemeCLIP for the hate detection task on PrideMM. We report macro-averaged Accuracy, AUROC, and F1-score to account for class imbalance. Each experiment is repeated with three different random seeds, and we report mean and standard deviation. These practices help minimize random variance and provide robust conclusions.

\paragraph{Expected Outcomes}
Given the demonstrated effectiveness of BLIP-2 and our use of residual adapter tuning, we expect our method (\textbf{MemeBLIP}) to outperform unimodal baselines and match or exceed state-of-the-art multimodal methods. Specifically, we anticipate better generalization on minority classes due to the semantic-aware design of our cosine classifier and improved modality alignment through independent projection layers.

\subsection{Baselines}

We compare our model MemeBLIP2 against several representative baselines across the unimodal and multimodal spectrum as well as the orginal MemeCLIP:

\begin{itemize}
    \item \textbf{BERT}: A fine-tuned BERT-base model trained on meme-extracted text.
    \item \textbf{GPT-4 (Text Only)}: Zero-shot inference using GPT-4o with prompt-based classification.
    \item \textbf{GPT-4 (Text + Image)}: GPT-4o with multi model input enabled.
    \item \textbf{MemeCLIP}: Trained and tested using the same settings as the original paper. 
    
\end{itemize}
These baselines enable a controlled comparison that isolates the contribution of multimodality and architectural enhancements.
\subsection{Main Results}

\begin{table*}[h]
\centering
\begin{tabular}{lccc}
\toprule
\textbf{Model} & \textbf{Accuracy (\%)} & \textbf{AUROC (\%)} & \textbf{F1-score (\%)} \\
\midrule
BERT & 70.8 & 70.8 & 70.2 \\
GPT-4o (Text Only) & 67.7 & 67.2 & 59.6 \\

GPT-4o (Text + Image) & 65.7 & 64.9 & 51.1 \\
MemeCLIP&74.4&\textbf{83.4}&73.0\\
\textbf{MemeBLIP-2 (Ours)} & \textbf{77.5} & {81.8} & \textbf{79.0} \\
\bottomrule
\end{tabular}
\caption{Classification performance on the PrideMM hate detection task. Results are reported as percentage values.}
\label{tab:main_results}
\end{table*}
 The PrideMM hate detection task receives performance results from multiple baselines and our proposed model through  Table~\ref{tab:main_results}.
 
We report classification performance using three metrics: accuracy (ACC), area under the ROC curve (AUROC), and macro-averaged F1 score. All experiments are repeated across three random seeds, and we report mean ± standard deviation.

The BERT model demonstrates the highest performance among  unimodal baselines by reaching 70.8\% accuracy and 70.2\%  F1-score while surpassing GPT-4 (67.7\% accuracy, 59.6\%  F1-score) and GPT-4o (65.7\% accuracy,  51.1\% F1-score) in this particular task. The results show that text-only transformers with  fine-tuning maintain strong performance in meme understanding compared to instruction-tuned models that lack multimodal adaptation.

 The proposed MemeBLIP-2 model demonstrates superior performance than all unimodal baselines by  reaching 77.5\% accuracy and achieving 81.8\% AUROC and  79.0\% F1-score. It also outperforms MemeCLIP, the paper that we mainly refer, by 3.1\% in accuracy and 6.0\% in F1-score, but the AUROC is slightly lower by  1.6\% .
 
 In a word, the proposed method of combining visual and textual information through effective alignment and fusion  proves beneficial for handling noisy meme content that contains rich contextual information.

\subsection{Gradient Diagnosis}

\begin{figure*}[t]
  \centering
  \includegraphics[width=\textwidth]{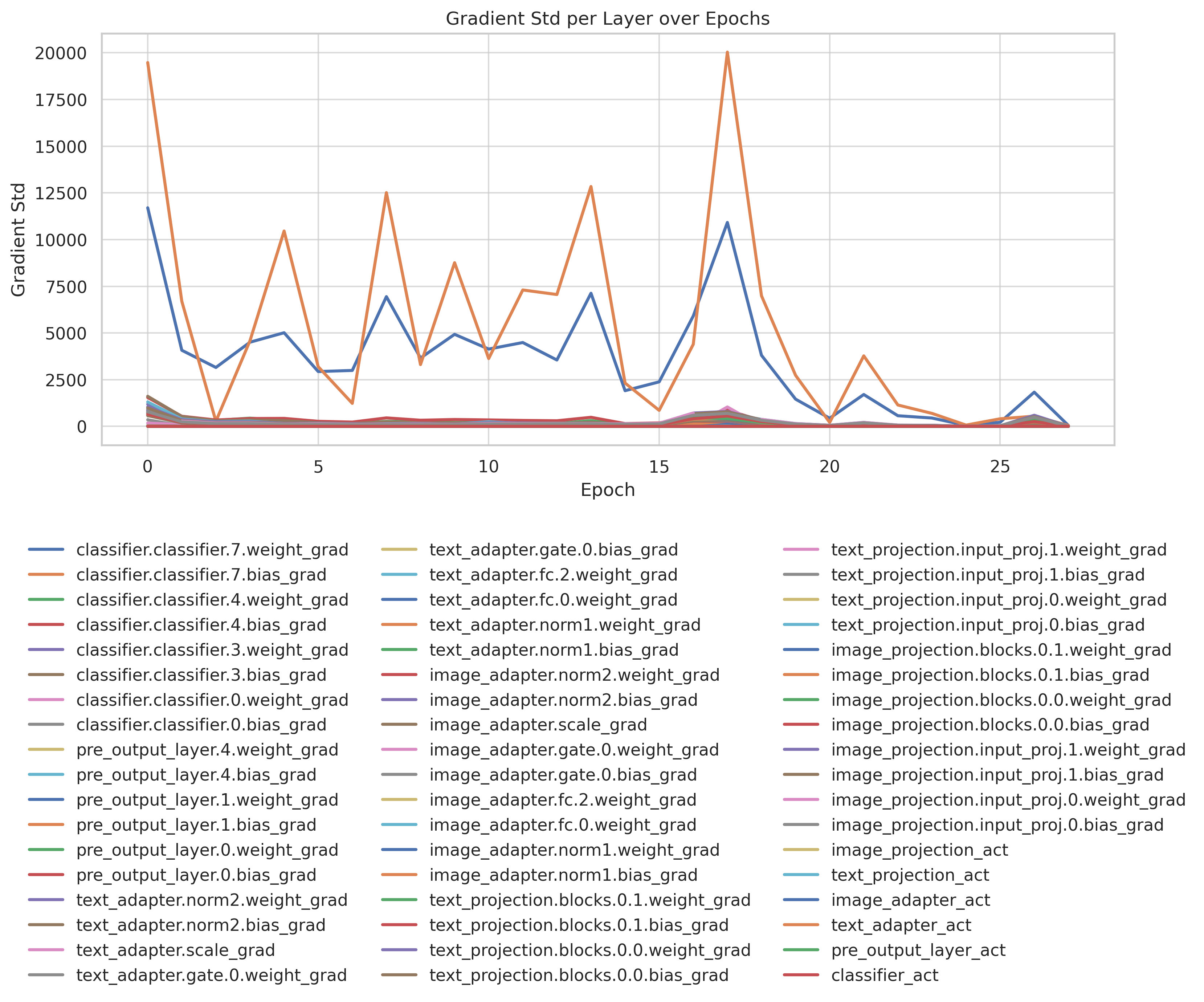}
  \caption{Gradient standard deviation of individual parameters across training epochs. Notable fluctuations are observed in the classifier’s final weight and bias terms.}
  \label{fig:grad-std}
\end{figure*}

\noindent
\noindent
To better understand the training dynamics of our model, we monitored the gradient standard deviation (\texttt{grad\_std}) of each parameter throughout training. Figure~\ref{fig:grad-std} shows that while most modules—such as the adapters and projection layers—maintain relatively stable gradient distributions, the classifier head exhibits significant variance. In particular, the \texttt{classifier.7.weight} and \texttt{classifier.7.bias} parameters display repeated spikes, exceeding 10,000 in gradient standard deviation during multiple epochs.

This behavior indicates heightened sensitivity or instability in the classifier head, which aligns with our earlier ablation findings: removing or simplifying the classifier leads to notable performance shifts. These fluctuations suggest that further regularization, layer-specific learning rates, or gradient clipping might be beneficial, especially for deeper or more nonlinear components such as MLP classifiers.

Overall, this analysis complements our architectural ablations by highlighting potential optimization bottlenecks and emphasizing the importance of balanced gradient flow across modular components.

\subsection{Ablation Study on PrideMM Dataset}

To assess the contribution of each module in our architecture, we conduct an ablation study on the PrideMM dataset using the element-wise multiplication (\textbf{mul}) fusion strategy. The results are shown in Table~\ref{tab:ablation-pridemm}.

The following configurations are evaluated:

\begin{itemize}
    \item \textbf{All modules (baseline)}: Full model using the MLP classifier, pre-output transformation layer, adapters, and image/text projection modules.
    \item \textbf{Remove MLP classifier}: Replace the MLP classifier with a simple linear layer without non-linear transformations.
    \item \textbf{Remove MLP + Pre-Output}: Additionally remove the pre-output layer and feed fused features directly into the classifier.
    \item \textbf{Remove MLP + Pre-Output + Adapter}: Further disable the adapter modules for both image and text modalities.
    \item \textbf{Remove All Modules}: Remove the classifier, pre-output layer, adapters, and both projection layers, reducing the model to a minimal structure.
\end{itemize}

\begin{table*}[t]
\centering
\caption{Ablation results on the PrideMM dataset using \textbf{mul fusion}. Metrics are reported as mean ± standard deviation (\%) over 3 runs.}
\label{tab:ablation-pridemm}
\begin{tabular}{lccc}
\toprule
\textbf{Scenario} & \textbf{ACC (\%)} & \textbf{AUROC (\%)} & \textbf{F1 (\%)} \\
\midrule
All modules (baseline) & \textbf{76.90 ± 0.55} & \textbf{80.80 ± 0.96} & 78.39 ± 0.61 \\
Remove MLP classifier & \textbf{76.90 ± 0.21} & 80.16 ± 0.81 & \textbf{78.67 ± 0.55} \\
w/o MLP + Pre-Output & 76.46 ± 0.41 & 79.28 ± 0.31 & 77.71 ± 0.82 \\
w/o MLP + Pre + Adapter & 75.44 ± 0.36 & 76.79 ± 3.90 & 77.25 ± 1.08 \\
Remove All Modules & 49.56 ± 4.07 & 48.22 ± 1.73 & 46.42 ± 9.14 \\
\bottomrule
\end{tabular}
\end{table*}

From Table~\ref{tab:ablation-pridemm}, we observe that the full model achieves the best AUROC and is among the best in accuracy and F1 score, indicating a strong balance across evaluation metrics. Removing the MLP classifier has minimal effect on accuracy, but slightly decreases AUROC, suggesting that the classifier architecture helps produce more calibrated predictions.

Further removing the pre-output layer causes a moderate drop across all metrics, particularly AUROC and F1. This implies the pre-output transformation contributes meaningfully to downstream decision quality. When adapters are also removed, the degradation becomes more evident, supporting their role in enhancing modality-specific representation learning. Finally, removing all major modules—including the projection layers—leads to a dramatic performance collapse, with accuracy dropping to 49.56\% and AUROC near random (48.22\%). This underscores the necessity of layered projections and modular components for effective multimodal integration.

\subsection{Analysis}

Our experiments confirm that unimodal baselines such as BERT and GPT-4 struggle with interpreting visually grounded irony and sarcasm in memes. This validates our hypothesis that effective multimodal integration is essential for nuanced meme classification.

\vspace{0.5em}
\noindent
While our model does not surpass MemeCLIP~\cite{Shah2024} in absolute performance, it provides an alternative architecture for multimodal fusion. Unlike MemeCLIP—which uses contrastively pre-aligned CLIP embeddings—our method builds on BLIP2, where vision and language features are extracted independently and not inherently aligned. This introduces fusion challenges, but also offers flexibility in modular design and encoder selection.

\vspace{0.5em}
\noindent
The ablation study (Table~\ref{tab:ablation-pridemm}) reveals several insights. When all modules are intact, the model achieves the highest AUROC (80.80\%) and maintains strong accuracy (76.90\%) and F1 (78.39\%). Removing only the MLP classifier has little impact on accuracy, but leads to a mild AUROC decline—indicating its role in shaping output calibration rather than feature quality. Further removal of the pre-output transformation causes additional degradation across metrics, particularly F1, showing its contribution to the final representation.

\vspace{0.5em}
\noindent
Removing adapter modules causes a larger performance drop, suggesting they play a meaningful role in modality-specific transformation. Finally, removing all major modules—including projection layers—leads to a performance collapse, confirming their critical role in aligning unaligned modalities. AUROC drops from 80.80\% to 48.22\%, close to random, and F1 drops by over 30\%.

\vspace{0.75em}
\noindent
\textbf{Key insights from the updated ablation experiments:}
\begin{itemize}
    \item \textbf{Projection layers} are indispensable for cross-modal alignment. Removing them causes near-random predictions (AUROC: 80.80\% $\rightarrow$ 48.22\%).
    \item \textbf{Adapters} help improve feature expressiveness and adaptation, contributing to stable and generalizable learning.
    \item \textbf{The MLP classifier} improves calibration and downstream discrimination, although its removal does not drastically harm performance.
    \item \textbf{The pre-output transformation} provides additional abstraction and improves F1, supporting its inclusion despite moderate standalone impact.
\end{itemize}

\begin{figure*}[ht]
    \centering
    \includegraphics[width=1\textwidth]{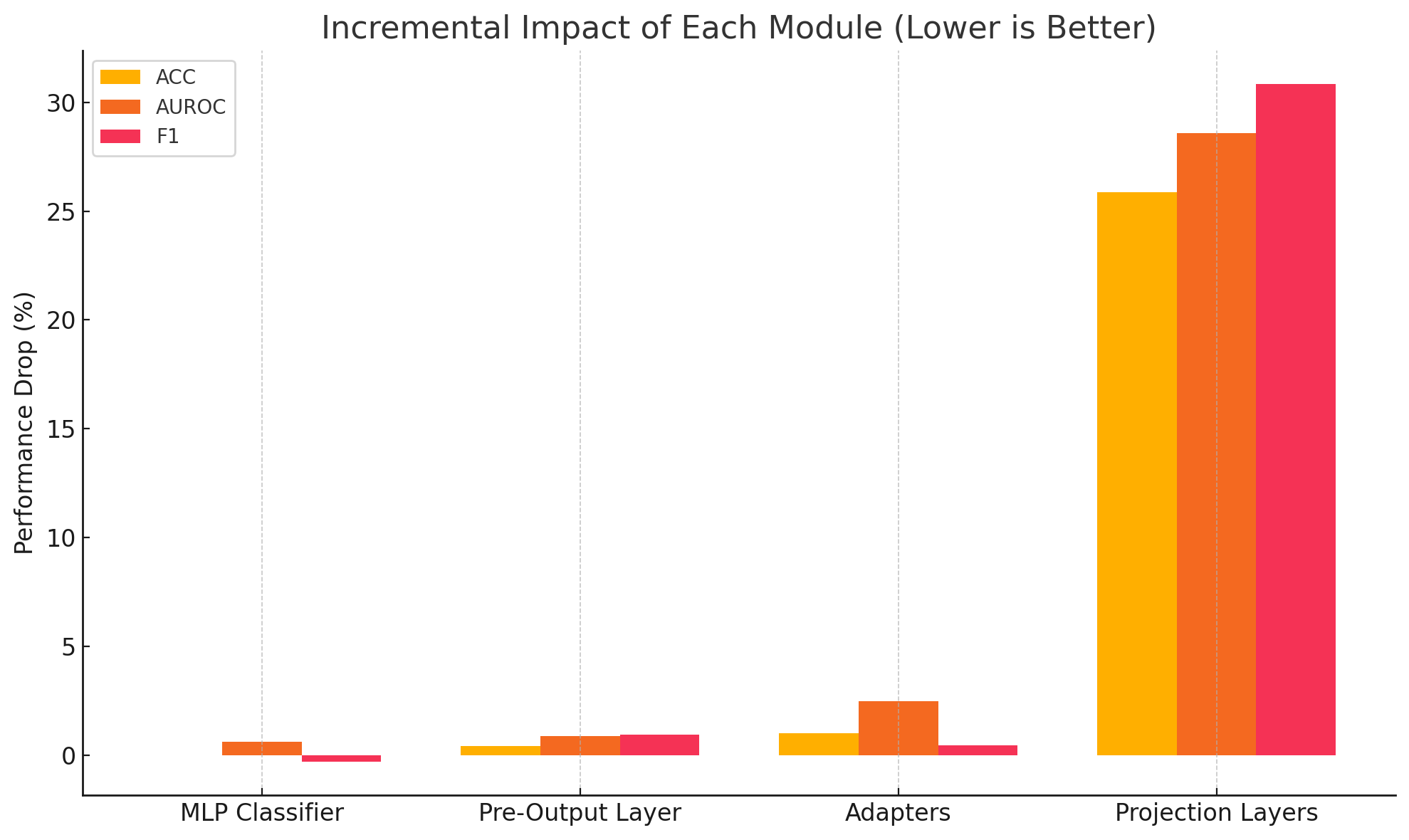}
    \caption{Incremental performance drop across metrics (ACC, AUROC, F1) when removing each module. Projection layers have the most significant impact, highlighting their role in modality alignment.}
    \label{fig:module-impact}
\end{figure*}

\vspace{0.75em}
\noindent
To better understand each component’s contribution, we visualize the performance degradation across accuracy, AUROC, and F1 score in Figure~\ref{fig:module-impact}. This incremental analysis confirms that while some modules (e.g., classifier, pre-output) offer moderate benefits, the adapter and projection layers play a pivotal role in preserving cross-modal semantics and stable training dynamics.

\vspace{0.5em}
\noindent
Taken together, our findings emphasize the importance of intermediate layers in bridging the semantic gap between vision and language features under non-contrastive, independently encoded settings. These results suggest that while partial simplifications may preserve some performance, the combination of modular design, MLP-based classification, and modality-specific transformations produces the most robust architecture for multimodal meme classification.

\section{Conclusion And Further Work}

In this paper, we proposed \textbf{MemeBLIP2}, a lightweight but effective multimodal system for the classification of harmful memes. By building upon MemeCLIP and integrating the BLIP-2 vision-language model, we introduced a modular architecture composed of modality-specific linear projection layers and adapter modules. These components align image and text representations into a shared semantic space while preserving computational efficiency. Evaluations on the PrideMM dataset demonstrate that MemeBLIP2 achieves competitive performance and excels in handling culturally nuanced and subtly ironic meme content. Our ablation studies underscore the critical role of the projection and adapter layers. The projection modules serve as semantic bridges between heterogeneous image and text embeddings, while the adapters refine feature representations through residual learning. The MLP-based classifier, equipped with nonlinear transformations, further enhances classification performance, particularly in scenarios involving class imbalance and complex semantic interactions. Unlike prior approaches that often rely on large-scale, resource-intensive architectures, MemeBLIP2 emphasizes architectural simplicity and efficiency. This design choice aligns with recent trends in developing compact multimodal models suitable for deployment in real-world scenarios, including environments with limited computational resources.
To enhance the deployability of MemeBLIP2 in real-time systems, future work will focus on model optimization techniques such as pruning, quantization, and knowledge distillation. These efforts aim to reduce inference latency and memory consumption, making the framework suitable for mobile and edge computing applications, including on-device content moderation. We also plan to broaden the applicability of MemeBLIP2 to multilingual and culturally diverse meme content. This includes integrating multilingual language models and expanding evaluation to cross-cultural datasets. Additionally, we intend to incorporate explainability mechanisms—such as attention visualization and gradient-based attribution—to improve the interpretability of the model’s decisions, which is crucial for transparency and ethical AI usage in content moderation.

In conclusion, MemeBLIP2 represents a step forward in efficient multimodal classification, offering a scalable, adaptable, and interpretable solution for harmful content detection. Its modular design and strong empirical performance provide a robust foundation for future advancements in multimodal understanding and ethical AI deployment.

\end{document}